\documentclass[12pt]{spieman}  
\usepackage{amsmath,amsfonts,amssymb}
\usepackage{graphicx}
\usepackage{setspace}
\usepackage{tocloft}
\usepackage{algorithmic}
\usepackage{textcomp}

\usepackage{amssymb}
\usepackage{amsmath}
\usepackage{booktabs}
\usepackage{multirow} 
\usepackage{blindtext}
\usepackage{microtype}
\usepackage{bm}

\usepackage{adjustbox}
\usepackage{colortbl}

\usepackage{array}
\newcolumntype{C}[1]{>{\centering\arraybackslash}p{#1}}

\def\etal{\emph{et al. }}

\usepackage{hyperref}
\hypersetup{
    colorlinks=true,
    linkcolor=blue,
    urlcolor=cyan,
    }
\usepackage{marginnote}

\newif\ifshowchangebar

\showchangebarfalse

\usepackage[table]{xcolor}

\title{Vector Field Attention for Deformable Image Registration}

\author[a]{Yihao~Liu}
\author[b]{Junyu~Chen,}
\author[a,c]{Lianrui~Zuo}
\author[a,*]{Aaron~Carass}
\author[a]{Jerry~L.~Prince}
\affil[a]{Department of Electrical and Computer Engineering, Johns Hopkins University, Baltimore, MD 21218}
\affil[b]{Department of Radiology and Radiological Science, Johns Hopkins School of Medicine, Baltimore, MD~21287}
\affil[c]{Laboratory of Behavioral Neuroscience, National Institute on Aging, National Institute of Health, Baltimore, MD~21214}

\cftpagenumbersoff{figure}
\cftpagenumbersoff{table} 
\begin{document} 
\maketitle

\begin{abstract}
Deformable image registration establishes non-linear spatial correspondences between fixed and moving images.
Deep learning-based deformable registration methods have been widely studied in recent years due to their speed advantage over traditional algorithms as well as their better accuracy.
Most existing deep learning-based methods require neural networks to encode location information in their feature maps and predict displacement or deformation fields though convolutional or fully connected layers from these high-dimensional feature maps.
In this work, we present Vector Field Attention~(VFA), a novel framework that enhances the efficiency of the existing network design by enabling direct retrieval of location correspondences.
VFA uses neural networks to extract multi-resolution feature maps from the fixed and moving images and then retrieves pixel-level correspondences based on feature similarity.
The retrieval is achieved with a novel attention module without the need of learnable parameters. 
VFA is trained end-to-end in either a supervised or unsupervised manner.
We evaluated VFA for intra- and inter-modality registration and for unsupervised and semi-supervised registration using public datasets, and we also evaluated it on the Learn2Reg challenge.
Experimental results demonstrate the superior performance of VFA compared to existing methods.
The source code of VFA is publicly available at \url{https://github.com/yihao6/vfa/}.
\end{abstract}


{\noindent \footnotesize\textbf{*}Aaron~Carass,  \linkable{aaron\_carass@jhu.edu} }

\begin{spacing}{2}   

\section{Introduction}
Deformable registration establishes non-linear spatial correspondences between a pair of fixed and moving images.
Traditionally, deformable registration is formulated as an energy minimization problem, in which the dissimilarity between the warped moving image and the fixed image and the irregularity of the deformation are jointly minimized for each individual pair of fixed and moving images.
Many successful algorithms, including LDDMM~\cite{beg2005computing}, SyN~\cite{avants2008symmetric}, and Elastix~\cite{klein2009elastix}, follow this approach. 

\begin{figure}[!t]
    \centering
    \includegraphics[width=0.65\textwidth]{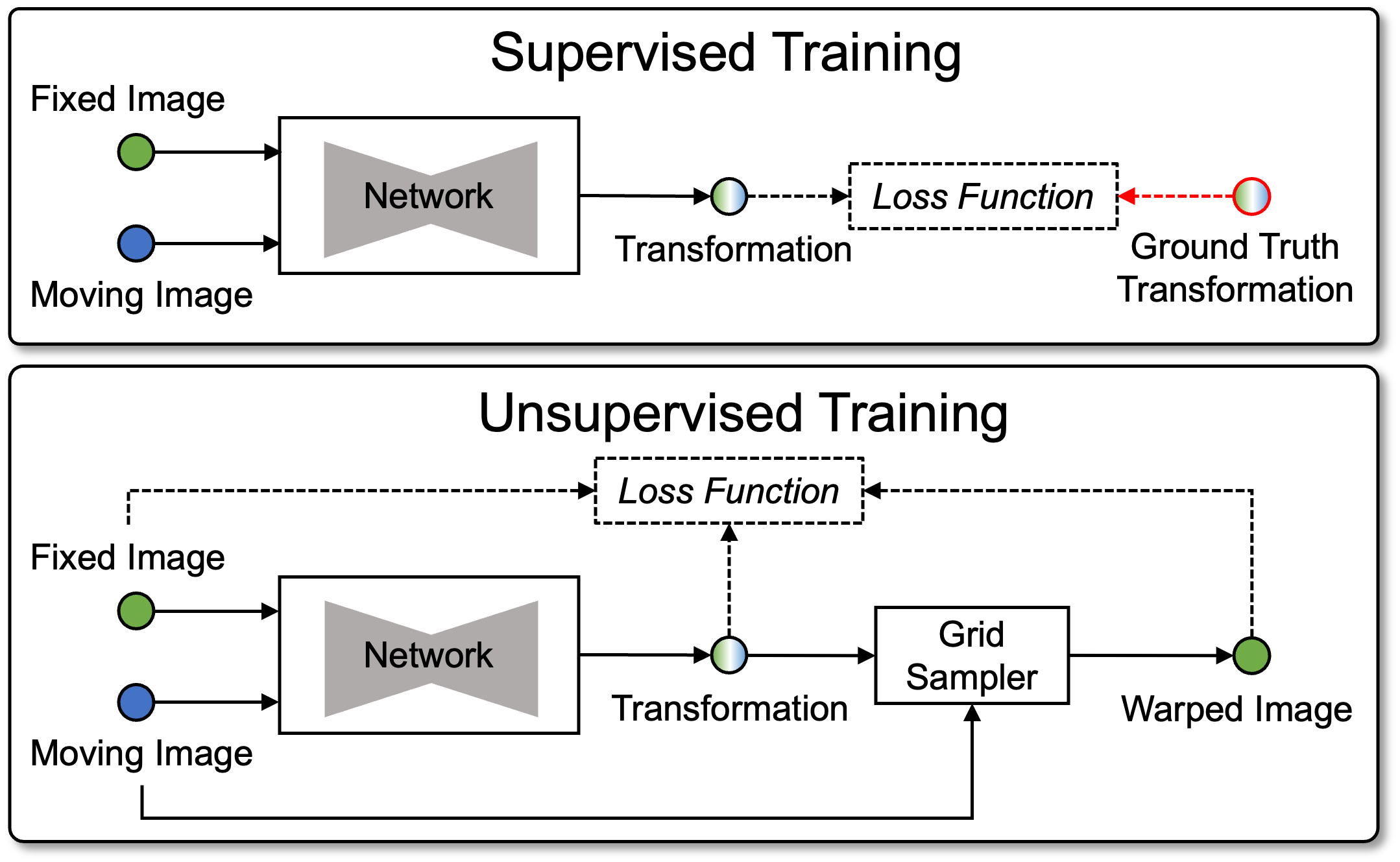}
    \caption{Overview of the supervised and unsupervised training schemes for deep learning based deformable registration algorithms.}
    \label{f:introduction}
\end{figure}

Deep learning-based methods take a fixed and a moving image pair as input to a neural network and give their spatial correspondence as output.
These methods are substantially faster to run because they avoid the pair-wise optimization process of conventional approaches by learning in advance a function that registers any pair of inputs at test time.
As depicted in Fig.~\ref{f:introduction}, there exist two training schemes for these deep learning-based methods~\cite{chen2023survey}.
Early methods required ground truth deformation for supervised training.
More recently, the integration of the differentiable grid sampler~\cite{jaderberg2015spatial} into the networks has enabled unsupervised training~\cite{de2017end}.
In both scenarios, the networks output the transformations represented by displacement or deformation fields through a convolutional or fully connected layer.
This approach yields higher registration accuracy when compared with traditional algorithms~\cite{de2017end, balakrishnan2019voxelmorph, chen2022transmorph, chenvit}.

However, the nature of the registration process requires neural networks to predict location correspondences with intensity images as inputs.
While deep learning has outperformed traditional hand-crafted methods in extracting features from these images, the tasks of feature matching and retrieval of correspondence from matched locations can effectively be handled by fixed operations, as demonstrated in classical algorithms~\cite{thirion1998image} and~\cite{shen2002hammer}.
Existing deep learning methods that rely on neural networks to predict a deformation field must not only learn to recognize and extract relevant features from the images, but also map those high-dimensional features back to spatial locations, through convolutional or fully connected layers.
We believe that such approaches dilute the effectiveness of neural networks.

To address this, we present vector field attention~(VFA), a novel deformable registration framework that enables direct location retrieval for producing a transformation.
VFA considers the registration task as a three-step process: feature extraction, feature matching, and location retrieval.
The \textit{feature extraction} step uses a feature extractor network to extract feature maps from the fixed and moving images independently.
In the \textit{feature matching} step, each integer-valued location in the fixed feature map is compared against the moving feature map at several candidate locations.
This results in an attention map, where those candidates that share similar features with the fixed location receive greater attention.
Our \textit{location retrieval} step retrieves the location of the candidates based directly on the feature similarity represented in the attention map. This yields the location correspondences.
What distinguishes our method, and in fact contributes significantly to its superior performance over existing algorithms, is the distinctive integration of the feature matching and location retrieval stages as fixed but differentiable operations.
Being fixed, they faithfully translate the knowledge of feature similarities to location correspondences, bypassing the need for learning to encode and decode location information.
The differentiability attribute ensures that the loss computed, whether in a supervised or unsupervised fashion, can be back-propagated, thereby enabling the feature extractor to learn and extract discriminative features that can create robust correspondences between the fixed and moving image.
We implement the feature matching and coordinate output steps together as a specialized attention module from the transformer architecture, with a vector field as one of the inputs.

VFA is an extension of our previously published method Im2grid~\cite{liu2022coordinate}, with improvements and extensive evaluations:
Firstly, we propose to replace the coordinate inputs in Im2grid with a more memory-efficient and flexible radial vector field.
Secondly, by carefully choosing the radial vector field to represent the relative displacements between voxels, we can remove the positional encoding layer used by Im2grid and rely on our specialized attention module to recover the locations for output.
Finally, we thoroughly test the proposed VFA on four different tasks:
1)~unsupervised atlas to subject registration of T1-weighted magnetic resonance~(MR) images; 
2)~unsupervised inter-modality T2-weighted to T1-weighted MR image registration;
3)~unsupervised inter-subject T1-weighted MR registration
from the Learn2Reg 2021 Challenge~\cite{hering2022learn2reg}; and
4)~semi-supervised intra-subject registration of inhale and exhale lung computed tomography~(CT) images from the Learn2Reg 2022 Challenge.
Our experiments demonstrate that the proposed method achieves state-of-the-art results in deformable image registration.

\section{Background and Related Works}

Denote the fixed and moving images as $I_f$ and $I_m$, respectively.
In this work, we consider two different \emph{digital} representations of a deformable transformation.
A transformation can be represented as a map of \emph{absolute} locations, denoted as $\phi$. For an integer-valued location $\bm{x}$, $\phi(\bm{x})$ represents the spatial correspondence between $\bm{x}$ in $I_f$ and $\phi(\bm{x})$ in $I_m$.
The intensity of the warped image $I_w$ at $\bm{x}$ can be easily acquired by sampling $I_m$ at $\phi(\bm{x})$.
In other words, the warped image can be written as $I_w(\bm{x}) = I_m(\phi(\bm{x}))$, which should be aligned with $I_f(\bm{x})$.
A deformable transformation can also be represented as a map of \emph{relative} locations known as a \emph{displacement field}~(denoted as $\bm{u}$).
The value of $\bm{u}(\bm{x})$ indicates the correspondence between $\bm{x}$ in $I_f$ and  $\bm{u}(\bm{x}) + \bm{x}$ in $I_m$.
The conversion between $\phi$ and $\bm{u}$ is achieved by
\begin{equation}
    \phi = \bm{u} + \phi_I,
    \label{e:disp_field}
\end{equation}
where $\phi_I$ is the identity grid that stores all integer-valued locations $\bm{x}$.
Both $\phi(\bm{x})$ and $\bm{u}(\bm{x})$ are defined for integer-valued locations, but they can take on floating-point values.
This is a ubiquitous strategy in deformable registration algorithms, as it offers convenience in rendering the warped image.
Otherwise, if correspondences were established between integer-valued locations in $I_m$ and floating-point locations in $I_f$, rendering the warped image would require interpolating scattered data points~\cite{crum2007methods, zhuang2008atlas}.

Deformable registration has traditionally been formulated as a minimization problem for each pair of fixed and moving image inputs. The function $L$ to be minimized takes the form of
\begin{equation}
    L(I_f, I_m, \phi) = \mathcal{L}_{\text{Sim}}(I_f, I_m\circ\phi) + \lambda\mathcal{L}_{\text{Reg}}(\phi),
    \label{e:energy}
\end{equation}
where $I_m\circ\phi$ denotes the application of $\phi$ on $I_m$, resulting in the warped image $I_w$.
The first term in Eq.~\ref{e:energy} penalizes the dissimilarity between $I_f$ and $I_w$, whereas the second term penalizes the irregularity of the transformation.
The hyper-parameter $\lambda$ controls the trade-off between these two terms.
Deep learning-based registration methods use neural networks to learn a generic function for registering two input images. Once trained, they can predict a transformation that aligns any two input images in a single forward pass.

In recent years, deep learning-based registration has witnessed notable advancement in both training strategy and network architecture.
Initial works trained convolutional networks in a supervised manner using the output from traditional algorithms~\cite{yang2017quicksilver, rohe2017svf} or artificial deformations~\cite{miao2016cnn,  eppenhof2018pulmonary}.
More recent works incorporate the differentiable grid sampler~\cite{jaderberg2015spatial} into their network structure.
This allows differentiable sampling of the moving image given the predicted transformation~\cite{de2017end}.
This enables unsupervised training of the network using a loss function similar to Eq.~\ref{e:energy}.

In terms of network structure, U-Net~\cite{ronneberger2015u} based architectures have emerged as the dominant choice for deformable image registration.
Several variants of the U-Net architecture have been proposed and achieved improved accuracy~\cite{balakrishnan2019voxelmorph, jia2022u, heinrich2019closing, heinrich2022voxelmorph++}.
To overcome the issue of the limited effective receptive field and better capture long-range spatial correspondences, Chen~\etal proposed a hybrid architecture that combines the Transformer structure with convolutional networks~\cite{chenvit, chen2022transmorph}.
The Transformer architecture adopts the attention mechanism, which is a differentiable operation that maps a query and a set of key-value pairs to an output.
It allows the model to selectively focus on different parts of the input when making predictions, based on their relevance to the task at hand.
The scaled dot-product attention used in the Transformer architecture is calculated as
\begin{equation}
    \text{Attention}(Q, K, V) = \text{Softmax}\left(
        \frac{QK^{T}}{\sqrt{d_k}}
    \right)V
    \label{e:attention}
\end{equation}
where $Q$, $K$, and $V$ are the matrices for queries, keys, and values; $\sqrt{d_k}$ is a scaling factor that depends on the dimension of $K$. The output of the $\text{Softmax}$ function is a matrix of attention weights, indicating the importance of each element in the key matrix for each element in the query matrix.
The attention weights are then used to compute a weighted sum of the corresponding values in the value matrix, producing the output of the attention mechanism.
In the Self-attention used in~\cite{chenvit, chen2022transmorph}, the feature maps of the fixed and moving images are concatenated and then used as input for the three components $Q$, $K$, and $V$.
This allows the model to selectively focus on different parts of the concatenated feature maps.
Further advancements for Transformer-based architectures in registration have been made by introducing the cross-attention modules~\cite{liu2022coordinate, song2022cross, shi2022xmorpher, chen2023deformable}.
In cross-attention, $Q$ is derived from the learned features of one input image, while $K$ and $V$ are derived from the learned features of the other input image. This allows the model to focus on different parts of the two input images and to capture relationships between them.

Few recent studies have delved into the inherent relationship between attention mechanisms and deformable registration.
In our prior work~\cite{liu2022coordinate}, we presented Coordinate Translator, which derived an attention map by comparing the fixed and moving image features.
This attention map is used to weight a map of coordinates within the moving image.
In~\cite{chen2022deformer}, the authors introduced Deformer, which employs concatenated feature maps of fixed and moving images to generate an attention map.
Contrary to the prevalent self-attention mechanism, their attention map is used to weight a three-channel feature map, which is interpreted as a map of basis displacement vectors.
Concurrent with this work, \cite{wang2023modet} developed ModeT, which extends the concepts of Im2grid and Deformer by incorporating multi-head attention.
However, both Deformer and ModeT still rely on convolutional layers to predict a deformation field.

\begin{figure*}[!t]
    \centering
    \includegraphics[width=0.99\linewidth]{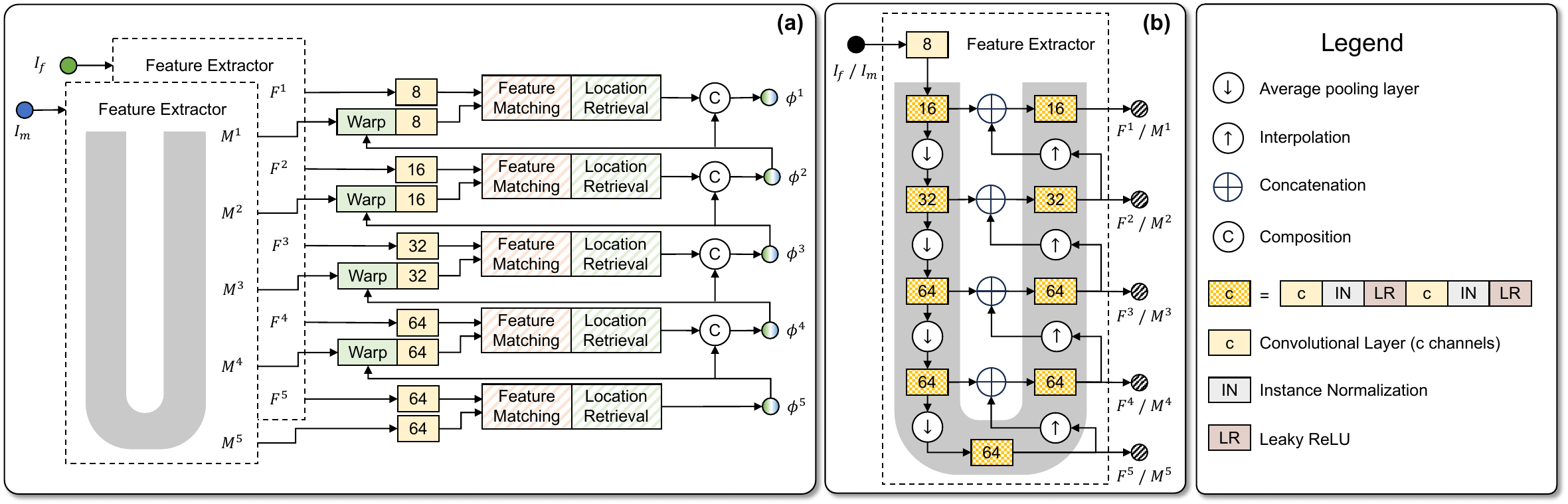}
    \caption{(a)~An overview of VFA and (b)~the detailed network architecture of the U-shape feature extractor network. The superscripts for the feature maps and transformations are used to indicate different spatial resolutions. $I_f$, $I_m$ denote the fixed, moving images.}
    \label{f:vfa}
\end{figure*}

\section{Method}
\noindent \textbf{Overview.}
VFA takes a fixed image $I_f$ and a moving image $I_m$ as inputs and produces $\phi^1$ as the final output to align the two images.
An overview of the VFA is depicted in Fig.~\ref{f:vfa}(a).
VFA considers the registration task as a three-step process: feature extraction, feature matching, and location retrieval.
The feature extraction step utilizes a feature extractor network to generate feature maps from the two input images at different resolutions.
At each resolution, feature matching and location retrieval steps utilize the extracted features to predict a transformation.
A multi-resolution strategy is adopted in which the extracted feature maps at finer resolution are pre-aligned using the transformation predicted from the previous coarser resolution.
The subsequent subsections provide the details of each component.

\subsection{Feature Extraction}
We use U-shaped networks to extract multi-resolution feature maps from $I_f$ and $I_m$, \emph{independently}.
The detailed structure of the feature extractor is shown in Fig.~\ref{f:vfa}(b).
We extract feature maps from $I_f$ at five resolutions, denoted as $F^1$, $F^2$, $F^3$, $F^4$, and $F^5$, and similarly extract $M^1$, $M^2$, $M^3$, $M^4$, and $M^5$ from $I_m$.
Larger superscripts indicate smaller spatial dimensions and lower resolutions.
For intra-modal registration, the two feature extractors share the same set of weights.
For inter-modal registration, we use a different set of weights for each feature extractor.

\begin{figure*}[!t]
    \centering
    \includegraphics[width=0.99\linewidth]{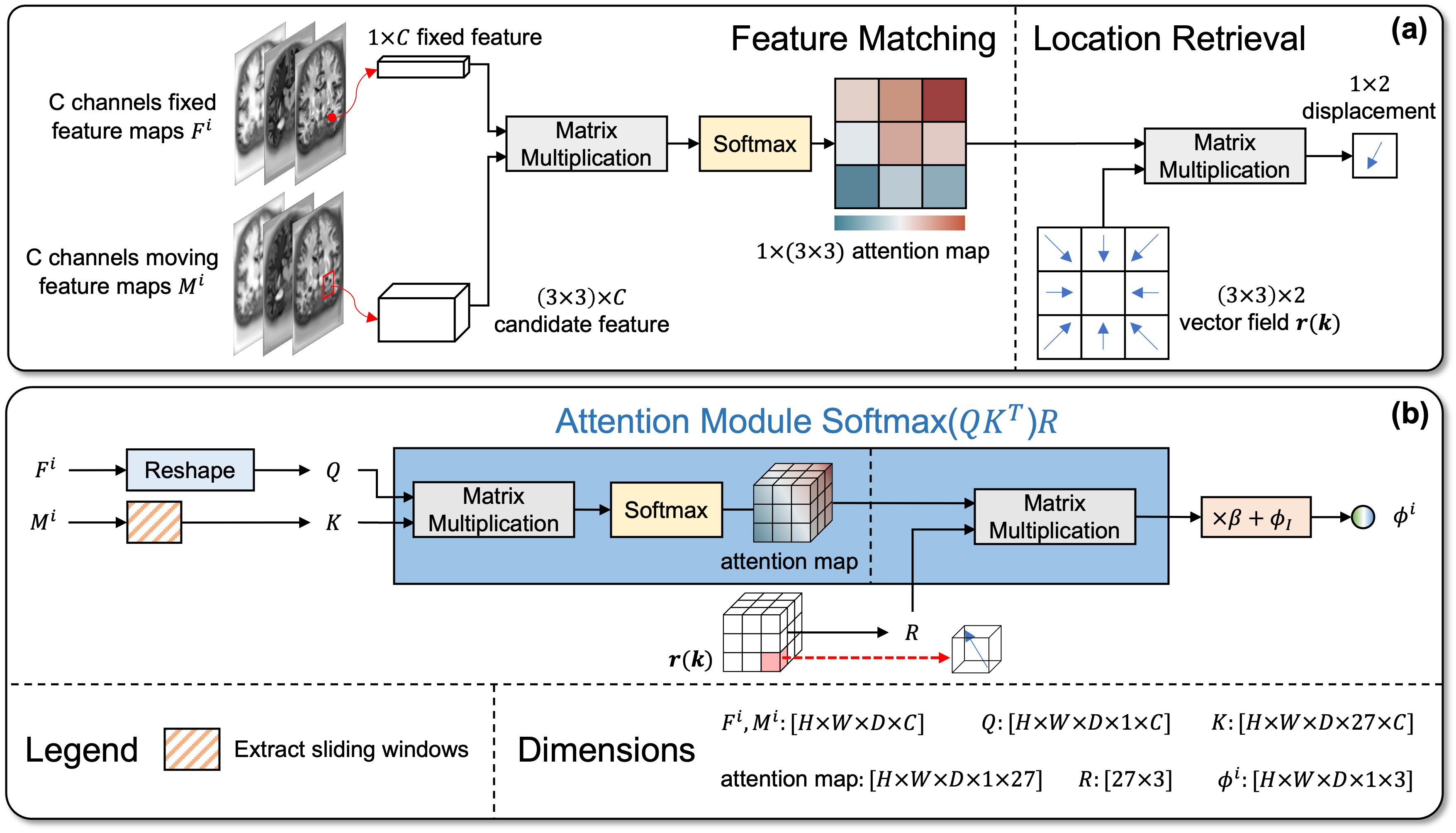}
    \caption{(a)~A 2D illustration of the feature matching and location retrieval steps for a single location in fixed feature maps. (b)~The 3D implementation of the feature matching and location retrieval steps using the specialized attention. The spatial dimensions are denoted as $H$, $W$, and $D$, respectively. The feature maps $F^i$ and $M^i$ are assumed to have $C$ channels.}
    \label{f:diagram}
\end{figure*}

\subsection{Feature Matching}
\label{ss:feature_matching}
Given feature maps $F^i$ and $M^i$ of the \emph{same resolution} with $C$ channels, extracted from $I_f$ and $I_m$ respectively, our feature matching step compares the feature $F^i(\bm{x})$ with a set of candidate locations in $M^i$.
In 2D, the candidate locations come from a $3\times 3$ search window centered at $\bm{x}$.
As illustrated in Fig.~\ref{f:diagram}(a), the $1\times C$ feature at $F^i(\bm{x})$ is compared with features from $M^i$ at nine candidate locations using the inner product.
The outputs are then normalized using a Softmax operation to create a $3\times3$ \emph{attention map}, indicating the order of candidate locations to find the best matches for $\bm{x}$ based on feature similarity.
Although, we only search for the best matches within the adjacent pixels of $\bm{x}$, long-range correspondences come from the use of our multi-resolution strategy.
For 3D images, $F^i(\bm{x})$ is matched with candidates in $M^i$ that are within the $3\times3\times3$ window centered at $\bm{x}$, resulting in a $3\times3\times3$ attention map.

We can efficiently implement the feature matching step using batched matrix multiplication~(which treats the last two dimensions as matrix dimension and other dimensions are broadcast).
Specifically, we construct a matrix $Q$ by reshaping $F^i$ and a matrix $K$ by extracting sliding windows of size $3\times3$~(2D) or $3\times3\times3$~(3D) from $M^i$.
The dimension of the matrices in our 3D implementation are shown in Fig.~\ref{f:diagram}(b).
The feature matching step is accomplished by computing $\text{Softmax}(QK^T)$, which exhibits a similar form to Eq.~\ref{e:attention}.

Note that our feature matching step is conceptually similar to the global correlation layer~\cite{heinrich2019miccai, xu2020cvpr, zhao2020cvpr} and local correlation layer~\cite{chen2023tmi, hansen2021tmi, kang2022mia, sun2018cvpr} used in previous works, which also generate an attention map. In these works, the attention map is processed through convolutional or fully connected layers to estimate a deformation field. However, as we will demonstrate, location correspondences are readily retrieved from the attention map with a fixed operation.

\subsection{Location Retrieval}
We now show that the location correspondences between $F^i$ and $M^i$ can be \emph{retrieved} from the attention maps $\text{Softmax}(QK^T)$ by completing the attention computation shown in Eq.~\ref{e:attention} with a carefully selected value matrix $V$.
In a typical application of attention, $V$ is the pool of features derived from the input images.
By weighting the features in $V$ according to the attention weights, the model can prioritize the most relevant ones for the task at hand.
In contrast, we define a vector field $\bm{r}(\bm{k})$ within the domain $\bm{k} \in \{-1, 0, 1\} \times \{-1, 0, 1\}$ for 2D, and $\bm{k} \in \{-1, 0, 1\} \times \{-1, 0, 1\} \times \{-1, 0, 1\}$ for 3D. Formally,
\begin{equation}
    \bm{r}(\bm{k}) = \bm{0}-\bm{k}.
\end{equation}
Consequently, each $\bm{r}(\bm{k})$ is a vector whose magnitude equals the Euclidean norm of $\bm{k}$, and is pointing to the origin of the vector field.
The only exception is at $\bm{r}(\bm{0})$, where it takes a zero vector.
Visual demonstrations of $\bm{r}(\bm{k})$ in 2D and 3D can be found in Fig.~\ref{f:diagram}.
The vectors defined in $\bm{r}(\bm{k})$ captures the displacement vectors between $\bm{x}$ and its adjacent candidates during the feature matching step.
To facilitate attention computation, all the vectors of $\bm{r}(\bm{k})$ are stored in a matrix $R$ of size $9\times2$~(2D) or $27\times3$~(3D), which serves as the value matrix.

This innovative form of attention allows us to prioritize the displacement vectors directly.
By computing the weighted sum of the vectors in $R$ using the attention of $\bm{x}$, we effectively retrieve the displacement of the candidates relative to $\bm{x}$ based on the feature similarity encoded in the attention map.
Since the attention map is soft, the output can be a \textit{floating-point displacement} rather than being limited to vectors within $\bm{r}(\bm{k})$.
This allows for more precise localization of correspondences.
For example, when the attention map of $F^i(\bm{x})$ highlights a single candidate in $M^i$, it retrieves the exact location of this candidate in the form of its displacement relative to $\bm{x}$.
When the attention map of $F^i(\bm{x})$ highlights multiple candidates in $M^i$~(as shown in Fig.~\ref{f:diagram}(a)), we obtain a weighted sum of the displacement vectors corresponding to those candidates.
The estimated displacement vector in the voxel coordinate system directly corresponds to a displacement in the scanner coordinate system in real-world units by applying the affine transformation that is associated with the volume data.
Although $R$ is fixed, our attention computation enables back-propagation of the loss.
Intuitively, to generate the desired displacements, the attention map must identify the correct set of displacements from $R$.
This process, in turn, encourages the feature extractors to learn to produce discriminative features that can yield robust correspondences between the fixed and moving images.

Overall, the feature matching and location retrieval steps are implemented as a specialized attention module that computes $\text{Softmax}(QK^T)R$, producing a displacement field $\bm{u}$.
In the final step, the network converts $\bm{u}$ to $\phi$, which can be used to warp images or feature maps using a grid sampler~\cite{jaderberg2015spatial}.
While it is straightforward to convert $\bm{u}$ to $\phi$ following Eq.~\ref{e:disp_field}, we add a \emph{learnable} parameter $\beta$ to the process:
\begin{equation}
    \phi = \beta\cdot\bm{u} + \phi_{I},
    \label{e:beta}
\end{equation}
where $\phi_{I}$ is the identity grid.
We introduce $\beta$ to ensure proper initialization of the training process.
Since the initial outputs of the feature extractors are not useful for registration, we set $\beta$ to a small number~(we used $\beta=0.1$ in all our experiments) to ensure that the initial output is close to $\phi_I$ instead of being completely random.
As the training progresses and the model learns to produce meaningful feature representations, we observe that $\beta$ automatically approaches a value near $1$.
Thus Eq.~\ref{e:beta} reduces to Eq.~\ref{e:disp_field}, indicating that the learned displacement field is fully incorporated into the final output.
However, it is noteworthy that the final
value to which $\beta$ converges can be influenced by both the initial $\beta$ value and the temperature parameter within the attention mechanism. Detailed experimental findings related to this are discussed in Section~\ref{s:beta}.

\begin{figure*}[!t]
    \centering
    \includegraphics[width=0.9\textwidth]{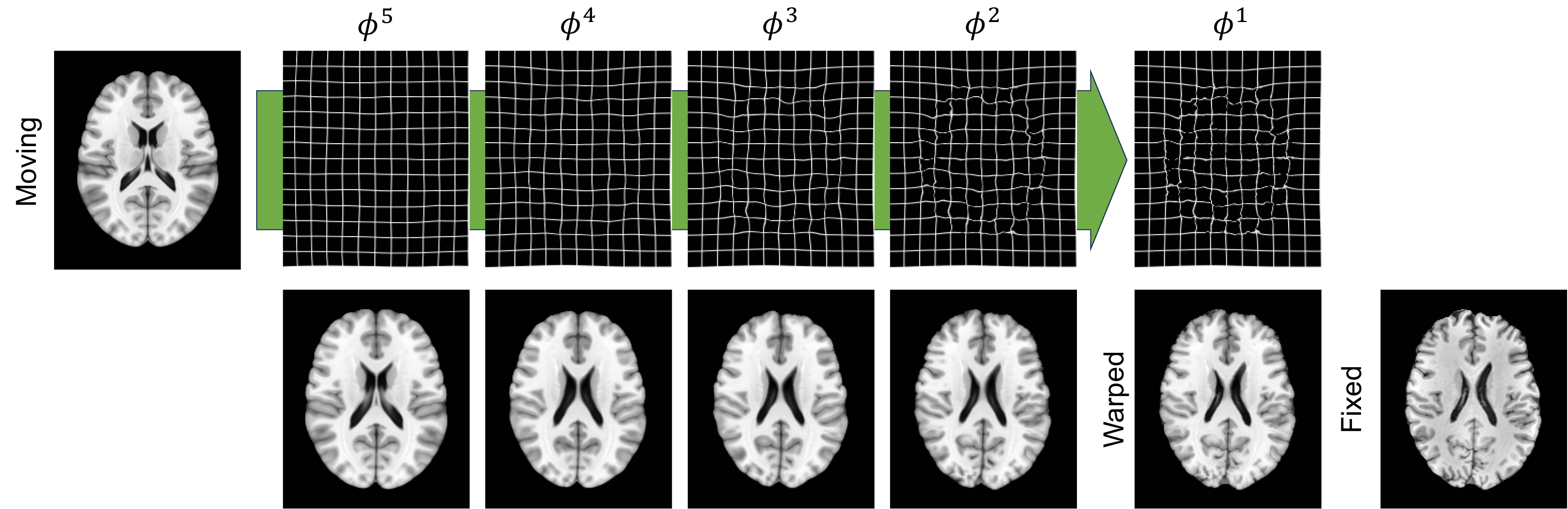}
    \caption{Visualization of the multi-resolution transformations.
    We used four downsampling steps in the feature extraction; Therefore, there are four intermediate low resolution transformations.
    For visualization purposes, these transformations have been upsampled to match the spatial dimensions of the input images.
    Additionally, each transformation has been applied to the moving image to visualize their effect.
    We note that only displacements within the axial plane are visualized in the grid line representations.
    In practical application, our algorithm outputs only the final transformation, $\phi^1$, and its corresponding warped image.}
    \label{f:multi_res}
\end{figure*}
\subsection{Multi-resolution Registration}
\label{s:vfa}
VFA deploys a multi-resolution strategy to register two images, as shown in Fig.~\ref{f:vfa}(a).
Once the multi-resolution feature maps are extracted, the feature matching and location retrieval steps are applied to the extracted features at each resolution.
To efficiently incorporate the resultant $\phi$'s at multiple resolutions, we use $\phi^{i+1}$ from the next coarsest resolution to warp the moving image features $M^i$, improving their alignment with $F^i$ before the feature matching, except for the lowest resolution where $M^4\circ \phi_{I}=M^4$.
This helps to resolve any coarse misalignments that may be present at lower resolutions.
An additional convolutional layer is applied to both $F^i$ and the warped $M^i$, respectively. 
It helps to rectify any distortions or inconsistencies in the feature space that might have been caused by the warping.
It also allows us to adjust the number of channels for the feature matching step to accommodate different memory budgets.
Because of the pre-alignment of $F^i$ and $M^i$, the feature matching and location retrieval steps produce a local transformation, which is then composed with $\phi^{i+1}$ to produce $\phi^i$.
Both the warping operation and the composition are accomplished by the differentiable grid sampler introduced in~\cite{jaderberg2015spatial}.
By using this coarse-to-fine strategy, we can capture large deformation with a small search window at each resolution. A visualization of the multi-resolution transformations can be found in Fig.~\ref{f:multi_res}. We note that the intermediate transformation~($\phi^5$ to $\phi^2$) are embedded in the network and their corresponding warped images shown in Fig.~\ref{f:multi_res} are only provided for illustration purposes.

The learnable parameter $\beta$, shared across all resolutions, is the only parameter that involves both the fixed and moving images; the remaining parameters in VFA only extract features from one of the input images, without the knowledge of the other input. This approach allows the learnable parameters of VFA to focus on learning generic features to recognize the input images, while leaving the feature matching and location retrieval to the specialized attention module.

In the unsupervised setting, the final output $\phi_{1}$ is applied to $I_m$ to produce the warped image $I_w$. This warped image is then used along with $\phi_1$ to compute the loss function given in Eq.~\ref{e:energy} during training.
In the supervised or semi-supervised setting,
the difference between $\phi_1$ and ground truth transformation can also be used as a loss function for network training.



\section{Experiments}
We implemented the proposed VFA using PyTorch.
In all experiments, we used the Adam optimizer with a learning rate of $1\times10^{-4}$ and a batch size of one for training.
The number of training epochs was determined using a validation dataset.
A random flip was applied to both input volumes simultaneously along the three axes as a data augmentation technique.
To evaluate the performance of VFA and demonstrate its versatility and robustness, we conducted experiments on four different tasks:
\begin{enumerate}
    \item unsupervised T1-weighted MR atlas to subject registration, in Section~\ref{s:exp_atlas};
    \item unsupervised multi-modal T2-weighted to T1-weighted MR registration, in Section~\ref{s:multimodal};
    \item unsupervised inter-subject T1-weighted MRI registration in the Learn2Reg 2021 Challenge~\cite{hering2022learn2reg} with scans from the OASIS dataset~\cite{lamontagne2019oasis}, in Section~\ref{s:unsupervised}; and
    \item semi-supervised intra-subject registration of inhale and exhale lung CT images in the Learn2Reg 2022 Challenge~\cite{hering2022learn2reg} with scans from the NLST dataset~\cite{national2011reduced}, in Section~\ref{s:ct-semisupervised}.
\end{enumerate}
The details for each of the tasks are provided in the subsequent sections.

\noindent \textbf{Evaluation Metrics.}
Due to the difficulty in acquiring manual landmark correspondences, for the T1-weighted~(T1w) MR atlas to subject registration and T2-weighted~(T2w) to T1w MR registration, we used the Dice similarity coefficient~(DSC) as a surrogate measure to evaluate the accuracy of the registration.
We first segmented all the scans using the deep learning-based whole-brain segmentation algorithm SLANT~\cite{huo20193d}.
We report the mean label DSC between the warped segmentation and fixed segmentation over the $132$ segmented labels.
To statistically evaluate the differences in DSC between VFA and each of the comparison methods, we employed the two-sided paired Wilcoxon signed-rank test~(null hypothesis: distribution of the DSC differences is symmetric about zero).
To measure the irregularity of the transformations, we report the number of non-diffeomorphic voxels~(ND~Voxels) computed using the central difference approximation of the Jacobian determinant. We also reported the non-diffeomorphic volume~(ND~Volume)~\cite{liu2022finite}, which measures the severity of the space folding under the digital diffeomorphism criteria.

For the unsupervised inter-subject T1w MR registration and semi-supervised intra-subject lung CT registration, we adopted the evaluation metrics used by the Learn2Reg Challenge.
For the inter-subject T1w MRI registration task, the segmentation of $35$ anatomic labels was acquired using FreeSurfer and SAMSEG from the neurite package~\cite{dalca2018anatomical}, and the accuracy was measured using the mean DSC and $95\%$ percentile of Hausdorff distance~(HD95) between the warped and fixed segmentations.
The accuracy of the intra-subject lung CT registration was measured by the target registration error~(TRE) using manual landmarks.
The $30\%$ lowest TRE~(TRE30) among all test cases is used to indicate the robustness of algorithms.
The smoothness of the transformations was evaluated using the standard deviation of the logarithm of the central difference approximated Jacobian determinant~(SDLogJ).

\noindent \textbf{Baseline Methods.}
In the first two experiments, we compared VFA with several state-of-the-art deep learning methods including: 1)~Voxelmorph~(VXM)~\cite{balakrishnan2019voxelmorph}: A deep learning method based on a U-Net architecture; 
2)~Voxelmorph-diff~(VXM-diff)~\cite{dalca2019unsupervised}: A variant of VoxelMorph with a scaling-and-squaring layer to encourage diffeomorphic registration; 
3)~TransMorph~\cite{chen2022transmorph}: A hybrid deep learning architecture that combines the Transformer structure with convolutional networks; 
4)~Im2grid~\cite{liu2022coordinate}: our previous method using coordinate translator modules;
5)~DMR~\cite{chen2022deformer}: A deep learning based architecture based on the Deformer module; and
6)~PR-Net++~\cite{kang2022mia}: A deep learning method that adopts a dual-stream architecture and incorporates correlation layers.
We did not include traditional registration methods in our experiments since the selected comparison methods have demonstrated superior performance over traditional methods in previous studies.

The experiments were conducted using NVIDIA RTX A6000 GPUs.
The number of training epochs in each experiment was determined using the validation dataset.
Typically, VFA achieves $95\%$ of its peak performance within 100,000 iterations.
At test time, VFA averages $0.9$ seconds per subject on our GPU, compared to $65.9$ seconds on a CPU at 2.61GHz.
All algorithms shows fast inference speeds, requiring only a few seconds. 
Given the minimal time differences, we did not focus on their comparison.

\subsection{Unsupervised T1w MR Atlas to Subject Registration}
\label{s:exp_atlas}
\noindent \textbf{Dataset.} We used the atlas image from~\cite{fonov-unbiased-2009} as the moving image and T1w scans from the publicly available IXI dataset~\cite{ixi} as the fixed images. For this experiment, $575$~T1w scans were divided into $400$ for training, $40$ for validation, and $135$ for testing.
All scans underwent N4 inhomogeneity correction~\cite{tustison2010n4itk} and were pre-aligned with the atlas image using a rigid transformation. A white matter peak normalization~\cite{reinhold2019evaluating} was applied to standardize the intensity scale.

\noindent \textbf{Implementation Details.}
The normalized cross correlation loss with a window size of $9\times9\times9$ was used for $\mathcal{L}_{\text{Sim}}$ and the diffusion regularizer~\cite{balakrishnan2019voxelmorph} for $\mathcal{L}_{\text{Reg}}$.
We set $\lambda = 1$ in Eq.~\ref{e:energy}, for all algorithms following the recommended value reported in~\cite{balakrishnan2019voxelmorph} and~\cite{chen2022transmorph}. For DMR, the extra losses computed using the intermediate displacement fields were also included. 

\begin{figure*}[!t]
    \centering
    \includegraphics[width = 0.95\textwidth]{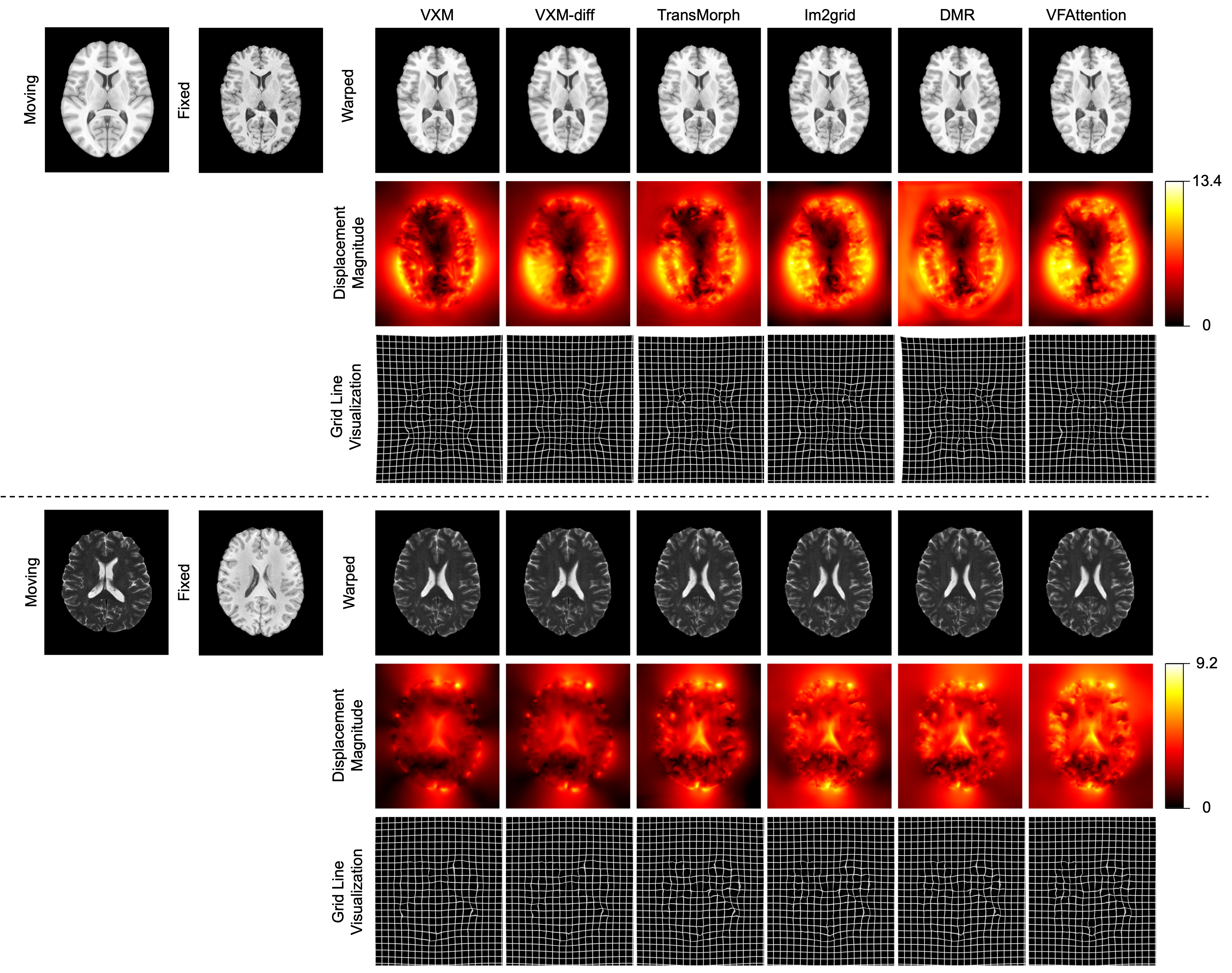}
    \caption{Visualization of the results for the T1w atlas to subject registration~(top) and the T2w to T1w registration~(bottom). The minimum and maximum values of the colorbar are specified in units of pixels.}
    \label{f:visualization}
\end{figure*}

\noindent \textbf{Results.} The performance of all the algorithms are summarized in Table~\ref{t:ixi_atlas}~(left).
VFA achieved the highest DSC among all algorithms with statistical significance~($\alpha=0.001$).
We also report the effect size, calculated between the proposed method VFA and the comparison method with the highest mean DSC~(Im2grid).
The result further reinforces the superiority of VFA. Specifically, the Rank-Biserial Correlation~(RBC)~\cite{kerby2014cp} was found to be $1.0$, indicating a perfect positive relationship in the differences between paired observations.
Additionally, the Common Language Effect Size~(CLES)~\cite{mcgraw1992pb} was found to be $0.9652$, suggesting that there is a $96.52\%$ chance that a randomly selected pair will exhibit a difference in the expected direction favoring VFA.
We also observed that VFA produced fewer folded voxels and smaller folded volumes compared with VXM, TransMorph, DMR, and PR-Net++ under the same choice of regularization weights $\lambda$.
This behavior is likely related to the local search strategy adopted in the feature matching step, as evidenced by the similar results produced by Im2grid, which utilized a similar strategy.
In contrast, TransMorph and DMR, which employed self-attention over a large window, did not exhibit this property.
The smoothness of the displacement fields produced by each algorithm can be observed in Fig.~\ref{f:visualization}.
We also implemented a variant of VFA~(\textit{VFA-Diff}) with the addition of the scaling-and-scaling-technique.
Both VXM-diff and VFA-diff demonstrated reduced folding compared to their original versions.
We note that the scaling and squaring layer can be incorporated in all the algorithms shown in Table~\ref{t:ixi_atlas}, although it cannot guarantee a perfect diffeomorphism due to the finite difference approximation of the Jacobian computation~\cite{liu2022finite}.

\begin{table*}[!t]
    \centering
    \caption{Results of the unsupervised registration from an atlas to T1-weighted MR images and from T2-weighted to T1-weighted MR images. The reported Dice similarity coefficient~(DSC) is the mean of $132$ labels segmented by SLANT. The number of non-diffeomorphic voxels~(ND Voxels) and the non-diffeomorphic volume~(ND Volume) were also included.
    The best performing algorithm in each column is bolded. The number of parameters is reported in units of a million~(M). The number of floating point operations~(FLOPs) is reported in units of a trillion~(T).
    \label{t:ixi_atlas}}
    \adjustbox{max width = 1.0 \textwidth}
    {{\renewcommand{\arraystretch}{1.3}
        \begin{tabular}{l C{0.15\textwidth} C{0.15\textwidth} C{0.15\textwidth}
        c
                          C{0.15\textwidth} C{0.15\textwidth} C{0.15\textwidth}
                          C{0.12\textwidth}
                          C{0.06\textwidth}
        }
            \toprule
            & \multicolumn{3}{c}{\textbf{T1w Atlas} $\longrightarrow$ \textbf{T1w Subject}}
            & \qquad &
            \multicolumn{3}{c}{\textbf{T2w Subject} $\longrightarrow$ \textbf{T1w Subject}}
            \\
            \cmidrule(lr){2-4} \cmidrule(lr){6-8}
            &
            \multirow{2}{*}{DSC $\uparrow$}
            & ND~Voxels $\downarrow$ & ND~Volume $\downarrow$
            &&
            \multirow{2}{*}{DSC $\uparrow$}
            & ND~Voxels $\downarrow$ & ND~Volume $\downarrow$ & Number of & \multirow{2}{*}{FLOPs}
            \\
            & & ($\%$) & ($\%$) && & ($\%$) & ($\%$) & Parameters &
            \\
            \cmidrule(lr){1-4}\cmidrule(lr){6-8} \cmidrule(lr){9-10}
            & &
            $19171\pm7111$ & $7280\pm2660$ && & $711\pm1359$ & $121\pm250$ &
            \\
            \multirow{-2}{*}{\textit{VXM}\cite{balakrishnan2019voxelmorph}}
            &
            \multirow{-2}{*}{$0.726\pm0.048$}
            & $(0.24\%)$ & $(0.09\%)$
            &&
            \multirow{-2}{*}{$0.593\pm0.053$}
            & $(0.01\%)$ & $(<0.01\%)$
            & \multirow{-2}{*}{0.3M} & \multirow{-2}{*}{0.61T}
            \\
            \rowcolor[gray]{.90}
            & & $\mathbf{0.9}\pm3.9$ & $\mathbf{0.4}\pm0.4$ && & $\mathbf{0.1}\pm0.3$ & $\mathbf{0.3}\pm1.7$ & &
            \\
            \rowcolor[gray]{.90}
            \multirow{-2}{*}{\textit{VXM-diff}\cite{dalca2019unsupervised}}
            &
            \multirow{-2}{*}{$0.713\pm0.037$}
            & $(<0.01\%)$ & $(<0.01\%)$
            &&
            \multirow{-2}{*}{$0.611\pm0.049$}
            & $(<0.01\%)$ & $(<0.01\%)$
            & \multirow{-2}{*}{0.3M} & \multirow{-2}{*}{0.61T}
            \\
            & & $23745\pm6749$ & $11743\pm2652$ && & $1303\pm2067$ & $205\pm349$ &
            \\
            \multirow{-2}{*}{\textit{TransMorph}\cite{chen2022transmorph}}
            &
            \multirow{-2}{*}{$0.774\pm0.029$}
            & $(0.3\%)$ & $(0.14\%)$
            &&
            \multirow{-2}{*}{$0.660\pm0.044$}
            & $(0.02\%)$ & $(<0.01\%)$
            & \multirow{-2}{*}{17.3M} & \multirow{-2}{*}{0.86T}
            \\
            \rowcolor[gray]{.90}
            & & $3136\pm2114$ & $1090\pm622$ && & $148\pm358$ & $26\pm32$ & &
            \\
            \rowcolor[gray]{.90}
            \multirow{-2}{*}{\textit{Im2grid}\cite{liu2022coordinate}}
            &
            \multirow{-2}{*}{$0.792\pm0.012$}
            & $(0.04\%)$ & $(0.01\%)$
            &&
            \multirow{-2}{*}{$0.668\pm0.025$}
            & $(<0.01\%)$ & $(<0.01\%)$
            & \multirow{-2}{*}{1.5M} & \multirow{-2}{*}{1.11T}
            \\
            & & $19431\pm6978$ & $7116\pm2297$ && & $952\pm1804$ & $145\pm300$ &
            \\
            \multirow{-2}{*}{\textit{DMR}\cite{chen2022deformer}}
            &
            \multirow{-2}{*}{$0.763\pm0.033$}
            & $(0.24\%)$ & $(0.09\%)$
            &&
            \multirow{-2}{*}{$0.671\pm0.038$}
            & $(0.01\%)$ & $(<0.01\%)$
            & \multirow{-2}{*}{8.0M} & \multirow{-2}{*}{4.92T}
            \\
            \rowcolor[gray]{.90} & & $10161\pm4124$ & $3707\pm1229$
            && & $85\pm76$ & $20\pm13$ & &
            \\
            \rowcolor[gray]{.90}
            \multirow{-2}{*}{\textit{PR-Net++}\cite{kang2022mia}}
            &
            \multirow{-2}{*}{$0.785\pm0.031$}
            & $(0.12\%)$ & $(0.04\%)$
            &&
            \multirow{-2}{*}{$0.650\pm0.066$}
            & $(<0.01\%)$ & $(<0.01\%)$
            & \multirow{-2}{*}{1.2M} & \multirow{-2}{*}{2.12T}
            \\
            \cmidrule(lr){1-4}\cmidrule(lr){6-8} \cmidrule(lr){9-10}
            & & $4556\pm2997$ & $1509\pm716$
            && & $11\pm19$ & $8\pm8$ &
            \\
            \multirow{-2}{*}{\textit{VFA}}
            &
            \multirow{-2}{*}{$\mathbf{0.806}\pm0.012$}
            & $(0.06\%)$ & $(0.01\%)$
            &&
            \multirow{-2}{*}{$\mathbf{0.725}\pm0.022$}
            & $(<0.01\%)$ & $(<0.01\%)$
            & \multirow{-2}{*}{3.7M} & 
            \multirow{-2}{*}{1.41T}
            \\
            \rowcolor[gray]{.90} & & $15\pm35$ & $15\pm6$
            && & $0\pm0$ & $0.004\pm0.011$ & &
            \\
            \rowcolor[gray]{.90}
            \multirow{-2}{*}{\textit{VFA-Diff}}
            &
            \multirow{-2}{*}{$0.801\pm0.013$}
            & $(<0.01\%)$ & $(<0.01\%)$
            &&
            \multirow{-2}{*}{$0.718\pm0.027$}
            & $(<0.01\%)$ & $(<0.01\%)$
            & \multirow{-2}{*}{3.7M} & 
            \multirow{-2}{*}{1.41T}
            \\
        \bottomrule
        \end{tabular}
    }}
\end{table*}

\subsection{Unsupervised T2w to T1w MR Registration}
\label{s:multimodal}
\noindent \textbf{Dataset.} We used the IXI dataset described in Sec.~\ref{s:exp_atlas}, with the same training, validation, and testing split.
Each training sample consists of a T2w scan as the moving image and a T1w scan as the fixed image. Both scans were selected at random from the training set.
We used the same preprocessing steps as used in Sec.~\ref{s:exp_atlas}, including inhomogeneity correction and rigid registration to an MNI space.
The intensity values of each image are normalized to the range $[0, 1]$ for both T2w and T1w images.
During validation and testing, we used a predefined set of $40$ and $135$ pairs of T2w and T1w scans. The two scans in each pair were selected from different subjects.

\noindent \textbf{Implementation Details.}
To account for the inter-modality registration task, we used the mutual information loss~\cite{guo2019multi} as $\mathcal{L}_{\text{Sim}}$ and diffusion regularizer~\cite{balakrishnan2019voxelmorph} as $\mathcal{L}_{\text{Reg}}$.
Since the comparison methods did not experiment with mutual information loss for inter-modality registration or provide a recommended value for $\lambda$, we set $\lambda=0.2$ for all algorithms such that the two losses were at similar scales during training.

\noindent \textbf{Results.} The performance of all the algorithms are summarized in Table~\ref{t:ixi_atlas}~(right).
Since inter-modality registration is a more challenging task than intra-modality registration, there is a decrease in registration accuracy for all algorithms.
Nevertheless, VFA achieved the highest DSC among all algorithms with statistical significance~($\alpha=0.001$). We also report the effect size, calculated between the proposed method VFA and the comparison method with the highest mean DSC~(DMR). The Rank-Biserial Correlation (RBC)~\cite{kerby2014cp} was found to be $0.9349$ and the Common Language Effect Size (CLES)~\cite{mcgraw1992pb} was found to be $0.9255$.
Sample results are shown in Fig.~\ref{f:visualization}.

%

\begin{figure*}[!t]
    \centering
    \begin{tabular}{c}
        \textbf{Inner Product}
        \\
        \includegraphics[width=0.95\textwidth]{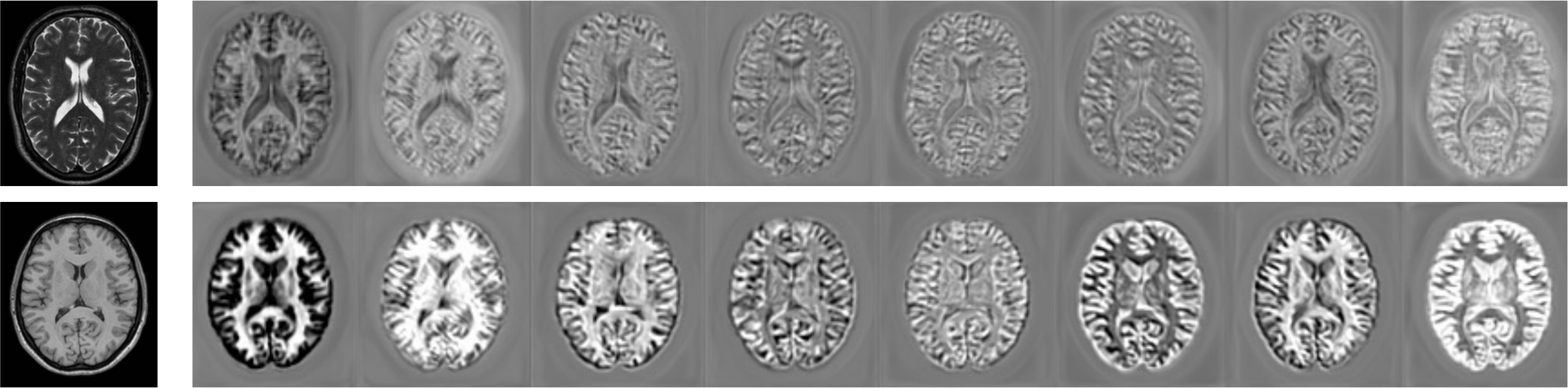}
        \\[0.5em]
        \textbf{Cosine Similarity}
        \\
        \includegraphics[width=0.95\textwidth]{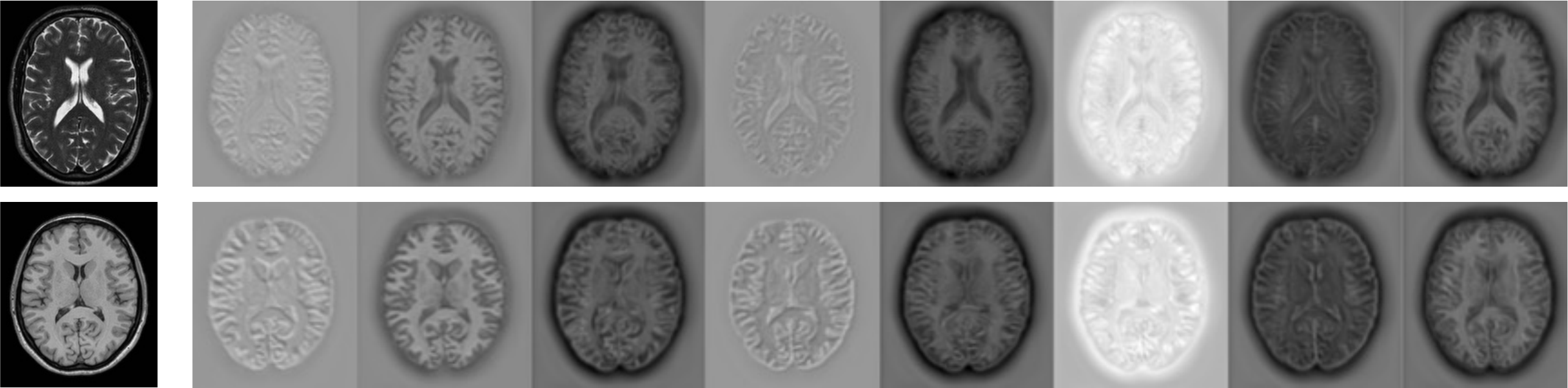}
    \end{tabular}
    \caption{Visualization of the feature maps extracted from the T2-weighted and T1-weighted images.
    Feature maps on the top are learned using inner product as the similarity in the attention computation.
    Feature maps on the bottom are learned using cosine similarity in the attention computation.}
    \label{f:feature}
\end{figure*}

Figure~\ref{f:feature} shows examples of the features extracted from T2-weighted and T1-weighted images prior to the feature matching step. Although the two images have different contrasts, the use of mutual information loss facilitates the learning of features that can be matched through inner product computation within the feature matching step.
However, a visual comparison of the corresponding features from the two modalities reveals a lack of high visual similarity.
We attribute this observation to two primary factors.
Firstly, our feature matching is localized, focusing on the highest similarity within small, defined areas and \emph{not} enforcing global similarity across the entire image.
Secondly and more critically, the inner product computation is sensitive to both the direction and magnitude of the feature vectors. 
Consequently, features deemed similar by the inner product may appear visually dissimilar due to variations in magnitude.
To verify this, we replaced the inner product with cosine similarity, which omits magnitude in the computation of similarity.
As illustrated in Fig.~\ref{f:feature}, using cosine similarity results in features that demonstrate substantially greater visual similarity.
In terms of performance, no noticeable difference in accuracy was observed; however, it is important to note that cosine similarity requires a slight increase in GPU memory usage.

\subsection{Perform Analysis on Model Capacity}
We acknowledge the significant impact of the number of learnable parameters on the performance of each model. Accordingly, we have detailed this information in Table I.
VXM and VXM-diff have fewer parameters due to their relatively small number of feature channels.
We conducted an extra comparison experiment where we doubled the number of feature channels across all convolutional layers in VXM, increasing its parameter count to 6.2~million. This high-capacity VXM model achieved a DSC of $0.771\pm0.030$ for T1w atlas to subject registration and $0.635\pm0.043$ for T2w to T1w registration.
Despite having considerably fewer parameters compared to this high-capacity VXM, as well as DMR and TransMorph, VFA demonstrates a statistically significant higher DSC.

To further investigate the effect of varying the feature extractor network on model accuracy, we evaluated two variants of VFA: 1)~VFA-Encoder, which uses the same encoder network as Im2grid; and 2)~VFA-Half, with the number of feature channels reduced by half compared to the original VFA.
VFA-Encoder achieved a DSC of $0.800\pm0.011$ for T1w atlas to subject registration and $0.696\pm0.023$ for T2w to T1w registration, both outperforming Im2grid.
VFA-Half achieved a DSC of $0.778\pm0.013$ for T1w atlas to subject registration---indicating a slight decrease in performance---yet it showed an improvement with a DSC of $0.729\pm0.019$ for T2w to T1w registration.

\begin{figure}[!t]
    \centering
    \includegraphics[width=0.7\textwidth]{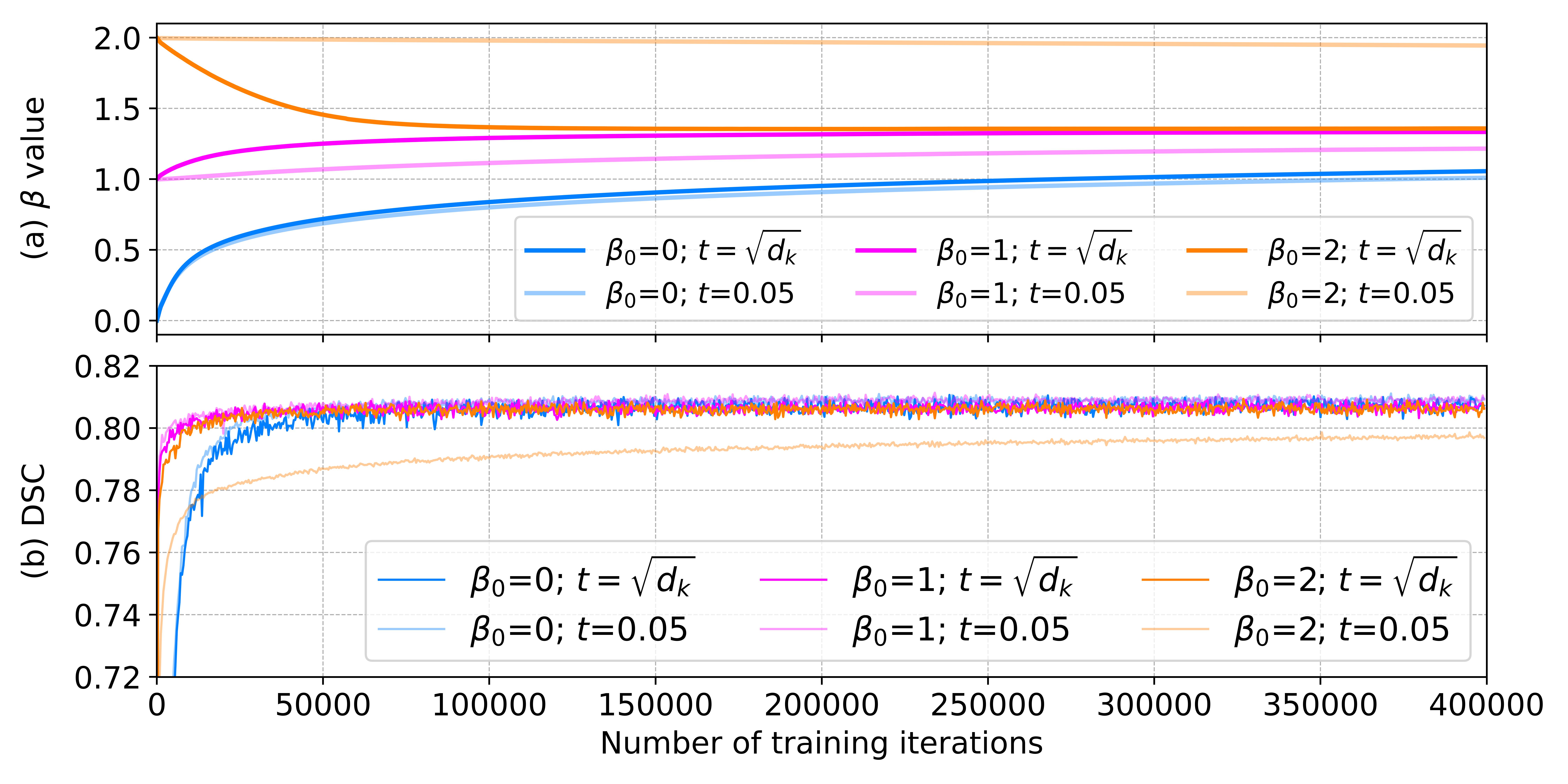}
    \caption{Illustration of $\beta$ values and model performance across training iterations. (a) shows the $\beta$ values for different initializations. (b) shows the corresponding Dice similarity coefficient.
    The notation $t=\sqrt{d_k}$ denotes the default temperature setting in scaled dot-product attention, while $t=0.05$ denotes the low temperature setting, designed to encourage a sparser attention.}
    \label{f:beta_plot}
\end{figure}

\subsection{Convergence properties of $\beta$}
\label{s:beta}
To ensure the registration starts from a reasonable initialization, we introduce a learnable parameter $\beta$ and set its initial value $\beta_0$ to $0.1$. In this section, we use the T1w MR Atlas to subject registration task to explore the impact of varying $\beta_0$ on model performance.
Figure~\ref{f:beta_plot}(a) shows the value of $\beta$ during training for initial values of $0$, $1$, and $2$. Figure~\ref{f:beta_plot}(b) shows the corresponding DSC observed on validation data throughout training.
When $\beta_0$ is initialized as $1$ or $2$, $\beta$ converges to approximately $1.3$ instead of $1$.
However, irrespective of the final beta values, all three models demonstrate similar performance.
This observation suggests different initialization can lead to a different level of sparsity in the attention map.
Specifically, when the attention weights are more evenly distributed, the weighted sum incorporates contributions from less similar points within the $3\times3\times3$ search window, effectively pulling the retrieved vector toward the center.
In these instances, $\beta$ converges to a value above $1$ to compensate for this effect.
To further verify this, we experimented with reducing the temperature parameter in the attention computation to $0.05$, aiming to encourage a sparse attention.
This adjustment leads to $\beta$ converging closer to $1$ then when $\beta_0$ is set to $0$ or $1$.
We note that a sparse attention can also present challenges in terms of optimization.
Particularly, starting the registration process with a $\beta_0$ of $2$, combined with a low temperature, adversely affects performance.

\begin{table}[!tb]
    \centering
    \caption{Results of the unsupervised inter-subject T1-weight MRI registration from Task~3 of the Learn2Reg 2021 Challenge~\cite{hering2022learn2reg}. The best DSC, HD95, and SDLogJ values among all methods are bolded. Standard deviations are not included because they were not reported in the original study~\cite{hering2022learn2reg}.
    \label{t:l2r2021}}
    \adjustbox{max width = 0.7 \linewidth}
    {{\renewcommand{\arraystretch}{1.3}
        \begin{tabular}{l C{0.15\textwidth} C{0.15\textwidth} C{0.15\textwidth}}
            \toprule
            &\multicolumn{3}{c}{\textbf{Learn2Reg 2021 Challenge Task3}}
            \\
            \cmidrule(lr){2-4}
            & DSC $\uparrow$ & HD95 $\downarrow$ & SDLogJ $\downarrow$
            \\
            \midrule
            \textit{Driver}~\cite{lv2022joint}
            &
            $0.799$
            &
            $1.775$
            &
            $0.079$
            \\
            \rowcolor[gray]{.90}\textit{ConvexAdam}~\cite{siebert2022fast}
            &
            $0.811$
            &
            $1.629$
            &
            $\mathbf{0.070}$
            \\
            \textit{LapIRN}~\cite{mok2020large}
            &
            $0.822$
            &
            $1.668$
            &
            $0.071$
            \\
            \rowcolor[gray]{.90}\textit{Im2grid}~\cite{liu2022coordinate}
            &
            $0.824$
            &
            $1.689$
            &
            $0.195$
            \\
            \textit{TransMorph}~\cite{chen2022transmorph}
            &
            $0.829$
            &
            $\mathbf{1.580}$
            &
            $0.093$
            \\ \hline
            \rowcolor[gray]{.90}\textit{VFA}
            &
            $\mathbf{0.834}$
            &
            $1.658$
            &
            $0.234$
            \\
        \bottomrule
        \end{tabular}
    }}
\end{table}

\subsection{Weakly-supervised inter-subject T1w MR registration}
\label{s:unsupervised}
\begin{figure}[!t]
    \centering
    \begin{tabular}{ccc}
        Moving & Warped & Fixed
        \\
        \includegraphics[width=0.2\textwidth]{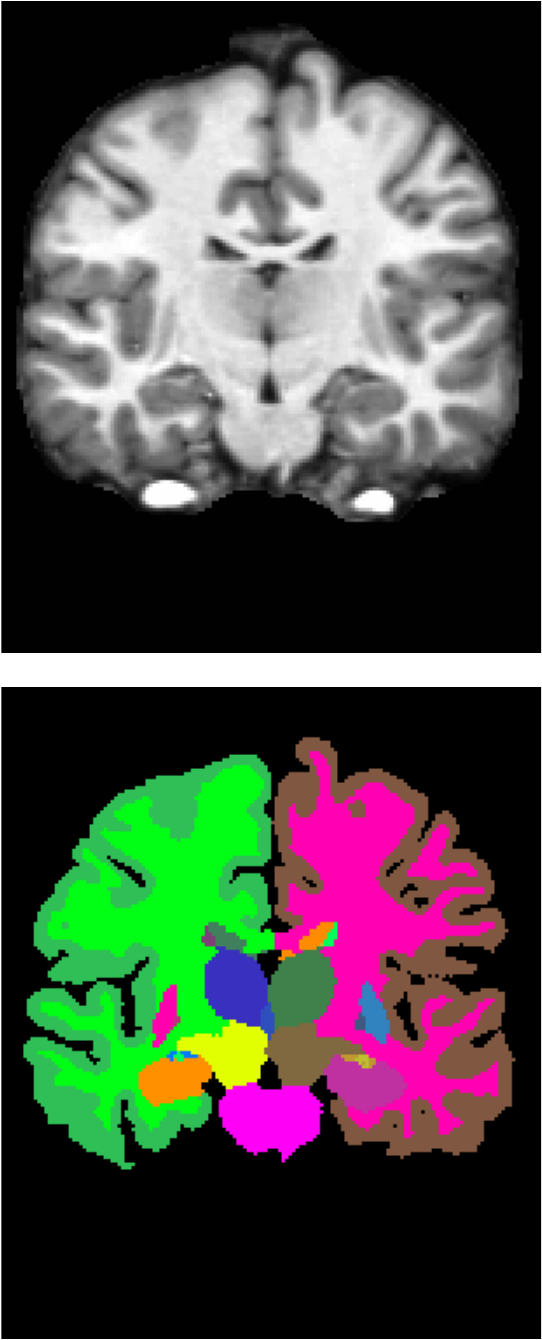}
        &
        \includegraphics[width=0.2\textwidth]{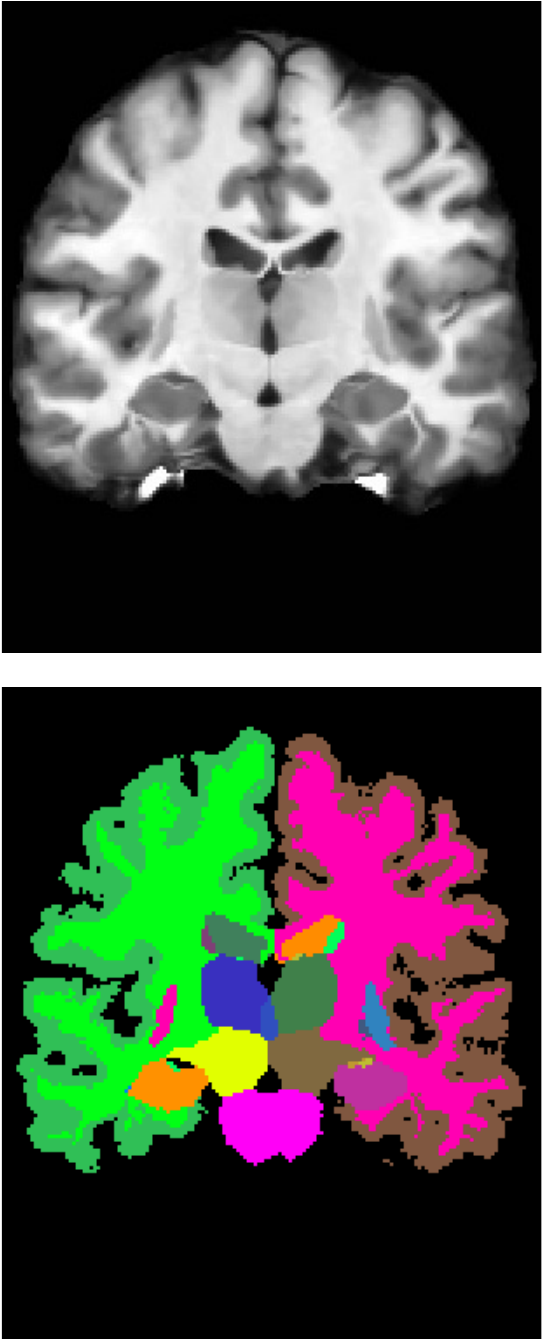}
        &
        \includegraphics[width=0.2\textwidth]{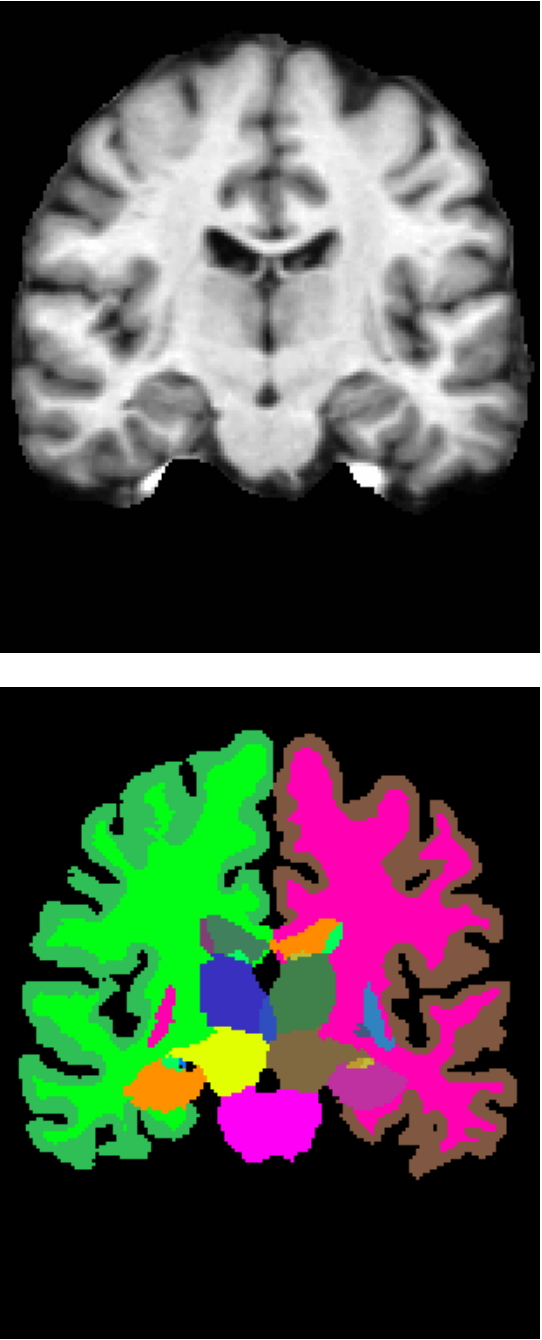}
    \end{tabular}
    \caption{Visualization of the results for the inter-subject brain MR image task from the Learn2Reg~2021 Challenge.
    The top row shows the intensity images, while the bottom row shows the corresponding Freesurfer labels.
    The moving and fixed labels are provided by the challenge organizer; the warped label is produced by applying the VFA transformation to the moving label image.}
    \label{f:oasis}
\end{figure}

We trained VFA using the $394$ training scans provided by the Learn2Reg Challenge~\cite{hering2022learn2reg}.
A combination of mean squared error loss, diffusion regularizer~\cite{balakrishnan2019voxelmorph}, and Dice loss was used, with the Dice loss incorporating auxiliary anatomical information from the provided segmentation map of each subject.
The weights for the losses~(mean squared error: 1; diffusion regularizer: 0.05; Dice: 1) were chosen following~\cite{jia2022u}.
The number of epochs was selected based on the best validation accuracy.
The performance of VFA on the test set, as well as the results of the top-performing methods from the challenge, were obtained from the challenge organizers and are presented in Table~\ref{t:l2r2021}.
We only included the top five algorithms based on the highest DSC values achieved. For a complete table including all submitted methods, interested readers can refer to~\cite{hering2022learn2reg}.
VFA achieved the highest DSC among all previous methods and ranked second in terms of HD95.
In terms of SDLogJ, VFA produced less smooth deformations, which could be related to the choice of weighting between the different losses during training and the errors in automatic segmentation maps.
Sample results are shown in Fig.~\ref{f:oasis}.

\begin{table}[!tb]
    \centering
    \caption{Results of the intra-subject lung CT registration from the Learn2Reg 2022 Challenge.
    We only included the top five algorithms based on the best TRE achieved as well as our previous method Im2grid.
    The best TRE, TRE30, and SDLogJ values among all methods are bolded. Standard deviations are not included because they were not reported by the challenge.
    \label{t:l2r2022}}
    \rowcolors{4}{cyan!10}{white}
    \adjustbox{max width = 1 \linewidth}
    {{\renewcommand{\arraystretch}{1.3}
        \begin{tabular}{l C{0.15\textwidth} C{0.15\textwidth} C{0.15\textwidth}}
            \toprule
            &\multicolumn{3}{c}{\textbf{Learn2Reg 2022 Challenge Task1}}
            \\
            \cmidrule(lr){2-4}
            & TRE $\downarrow$ & TRE30 $\downarrow$ & SDLogJ $\downarrow$
            \\
            \midrule
            \textit{Im2grid}~\cite{liu2022coordinate}
            &
            $2.243$
            &
            $3.054$
            &
            $0.085$
            \\
            \textit{ConvexAdam}~\cite{siebert2022fast}
            &
            $2.212$
            &
            $3.617$
            &
            $\mathbf{0.035}$
            \\
            \textit{corrfield}~\cite{heinrich2015estimating}
            &
            $1.830$
            &
            $2.633$
            &
            $0.160$
            \\
            \textit{VROC}
            &
            $1.763$
            &
            $2.409$
            &
            $0.037$
            \\
            \textit{LKU-Net}~\cite{jia2022u}
            &
            $1.699$
            &
            $2.409$
            &
            $0.052$
            \\
            \textit{cwmokab}
            &
            $\mathbf{1.673}$
            &
            $2.394$
            &
            $0.044$
            \\ \hline
            \textit{VFA}
            &
            $1.705$
            &
            $\mathbf{2.311}$
            &
            $0.064$
            \\
        \bottomrule
        \end{tabular}
    }}
\end{table}

\subsection{Semi-supervised intra-subject lung CT registration}
\label{s:ct-semisupervised}
\begin{figure}[!t]
    \centering
    \begin{tabular}{ccc}
        Moving & Warped & Fixed
        \\
         \includegraphics[width=0.30\textwidth]{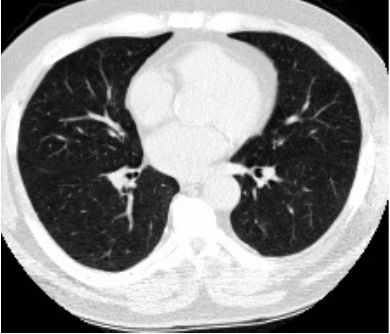}
         &
         \includegraphics[width=0.30\textwidth]{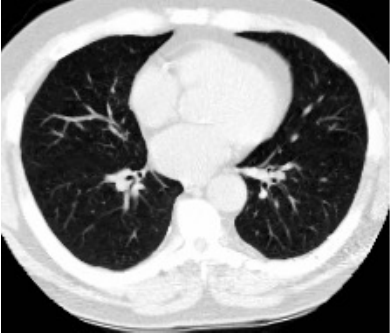}
         &
         \includegraphics[width=0.30\textwidth]{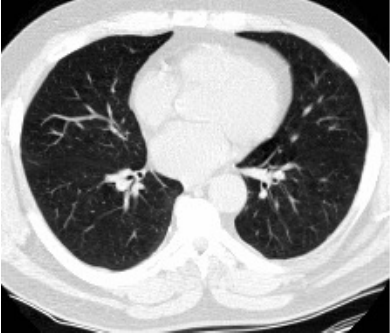}
    \end{tabular}
    \caption{Visualization of the results for inhale and exhale lung CT registration task from the Learn2Reg~2022 Challenge.}
    \label{f:nlst}
\end{figure}

We divided the 100 training pairs provided by the challenge into a $9:1$ for training and validation.
A combination of mean squared error loss, diffusion regularizer~\cite{balakrishnan2019voxelmorph}, and target registration error~(TRE) loss was used.
The TRE loss was computed from $\phi$ at locations where automatically generated keypoints~\cite{heinrich2015estimating} were provided by the challenge. Therefore, VFA was trained in a semi-supervised manner.
We empirically chose the weights for the three losses to be $5, 0.2, 0.05$.
We did not train with only TRE loss because the landmark correspondences in the test set were manually acquired, and thus, are different from the automatically generated keypoint correspondences.
The number of epochs was selected based on the best validation accuracy.
The performance of VFA on the test set, as well as the results of the top-performing teams from the challenge, were obtained from the challenge organizers and are presented in Table~\ref{t:l2r2022}.
VFA ranked among the top three with the lowest TRE. It also achieved the best TRE30, demonstrating superior robustness compared to all other methods. Sample results are shown in Fig.~\ref{f:nlst}.

\section{Conclusion and Discussion}
\label{sec:discussion}
In this paper, we proposed a deep learning-based deformable registration method called VFA.
It utilizes a novel specialized attention mechanism for feature matching and location retrieval, enabling the registration of intra-modal and inter-modal medical images with high accuracy.
Our experimental results show that VFA outperforms state-of-the-art registration methods on various datasets.

The maximum deformation achievable by VFA is inherently limited by its search region during the feature matching.
While this might seem like a limitation, it acts as a safeguard against deformation hallucination because location correspondences are established based solely on matched features.
Unlike VFA, methods that infer a deformation field via convolutional or fully connected layers have the potential to generate any deformations within the data type's range, which can extend beyond the network's receptive field.
The search region of VFA is directly influenced by the number of resolution levels it employs.
At its lowest resolution, VFA assigns a $3\times3\times3$ search window to each voxel.
Each subsequent upsample operation followed by composing with the transformation at higher resolutions effectively doubles the existing search region and then adds an additional two units to each dimension.
As a result, employing a five-level network allows VFA to attain a search region of $78\times78\times78$ for each voxel.
This covers approximately one-third of the voxel count along any axis in our applications.
However, when the number of resolution levels is reduced by one, the search region is reduced to a $38\times 38 \times 38$.
This adjustment was observed to significantly impair performance in the Lung CT registration tasks, where the variation between inhale and exhale CT scans demands estimating larger deformations.

One limitation of VFA is its memory usage, which is a common challenge for networks based on scaled dot-product attention. This issue is partly attributed to the use of high-resolution images in our experiments. As a result, we were unable to experiment with larger search windows than $3 \times 3 \times 3$. While a larger window, such as $5 \times 5 \times 5$, might offer potential improvements, we were constrained in our ability to test this hypothesis.
One possible solution to this issue is to employ more efficient attention mechanisms, such as those based on sparse attention~\cite{kitaev2020reformer} or learned sampling~\cite{xia2022vision}.
However, these approaches may come at the cost of reduced accuracy, and their effectiveness in the context of deformable registration remains an area for future research.

An important direction for future work is to extend VFA to handle time-series data such as 4D MRI or CT~\cite{bian2023midl, yu2023spie}.
VFA, which separates feature extraction from feature matching and location retrieval, is well-suited for handling 4D registration efficiently.
This is because the feature representations need only be computed once for each time point and can be reused for registering scans between any two time points.
This offers benefits in two aspects. 
Firstly, we can easily incorporate scans acquired across different time intervals to impose extra constraints during training and improve the overall accuracy without the need to rerun the entire network.
Secondly, the trained model can be flexibly adopted to register any pair of images in the time series.

There are several other promising directions for the application of the proposed VFA. One potential area of interest is in the field of multi-atlas segmentation of brain MR images.
While deformable registration is a crucial component of multi-atlas segmentation, existing registration methods are often time-consuming due to the need for multiple registrations and can suffer from a lack of accuracy.
VFA's ability to perform fast and accurate registration could potentially be beneficial for multi-atlas segmentation by reducing the time required for the registration step and improving the accuracy of deformable registration.
This could subsequently enhance the quality of multi-atlas segmentation.
Furthermore, the ability of VFA to handle inter-modal registration could be especially useful in cases where the atlases and target images are acquired using different modalities.
Finally, while the proposed method is designed for medical image registration, its specialized attention mechanism could potentially be applied to other computer vision tasks that involve feature matching and location retrieval, such as object detection or optical flow estimation.

\subsection*{Ethics Statement}
The IXI dataset was approved by the Institutional Review Board~(IRB)
of Imperial College London, in conjunction with the IRBs of
Hammersmith Hospital, Guy's Hospital, and the Institute of Psychiatry
at King's College London. The OASIS dataset is an open-access database
which had all participants provide written informed consent to
participate in their study. All OASIS participants were consented into
the Charles F. and Joanne Knight Alzheimer Disease Research Center
following procedures approved by the IRB of Washington University
School of Medicine. The National Lung Screening Trial~(NLST) recruited
potential participants and evaluated their eligibility for their
clinical trial. Individuals who were ruled eligible, signed an
informed consent form. A National Cancer Institute IRB reviewed the
consent forms and approved the study.

\subsection*{Disclosures}
The authors have no relevant financial interests and no other potential conflicts of interest to disclose
that are relevant to the content of this article.

We have used AI~(OpenAI GPT-4o) as a tool in the creation of this content, however, the foundational ideas, underlying concepts, and original gist stem directly from the personal insights, creativity, and intellectual effort of the author(s). The use of generative AI serves to enhance and support the author's original contributions by assisting in the ideation, drafting, and refinement processes. All AI-assisted content has been carefully reviewed, edited, and approved by the author(s) to ensure it aligns with the intended message, values, and creativity of the work.

\subsection* {Code, Data, and Materials Availability} 
The datasets used in this work are available in the IXI repository, \url{https://brain-development.org/ixi-dataset/}. The datasets related to the Learn2Reg challenge can be found at \url{https://learn2reg.grand-challenge.org}.
The source code of VFA is publicly available at \url{https://github.com/yihao6/vfa/}.
Links to the Docker container, Singularity Container, and Pretrained models of VFA are also available on the github page.

\subsection* {Acknowledgments}
This work was supported in part by the NIH/NEI grant R01-EY032284, the NIH/NINDS grant R01-NS082347, and the Intramural Research Program of the NIH, National Institute on Aging. Junyu~Chen was supported by grants from the NIH, U01-CA140204, R01-EB031023, and U01-EB031798.
The work was made possible in part by the Johns Hopkins University Discovery Grant~(Co-PI: J.~Chen, Co-PI: A.~Carass).
We are grateful to Dr.~Yong~Du for the generous grant support, which enabled 
the progression and completion of the research reported in this paper.
We would like to acknowledge the organizers of the Learn2Reg Challenge.
Their efforts have enabled the development of our proposed method, and contributed to the advancement of the field of medical image registration. 
Special thanks are extended to Christoph Grossbroehmer for his invaluable assistance in evaluating our algorithm.


\bibliography{report}   
\bibliographystyle{spiejour}   


\listoffigures
\listoftables

\end{spacing}
\end{document}